# Detection and Interpretability Analysis of Quotation Errors by Large Language Models

Bei Huang[1], Yingyi Zhang[1,*], Shenghao Huang[1], Chengzhi Zhang[2]

[1] School of Social Science, Soochow University, Soochow, China

[2] School of Economics and Management, Nanjing University of Science and Technology, Nanjing, China

**Abstract**

**Purpose -** Quotation error refers to the inconsistency between cited information and its original source. This phenomenon leads to a series of negative impacts, such as misinterpretation of the original research, undermining the academic community's collective understanding of relevant issues, and weakening the accuracy and fairness of the citation-based academic evaluation system. Existing studies have shown that quotation error is prevalent in the academic community; moreover, manual verification of quotation error is not only labor-intensive but also inefficient. Therefore, this paper proposes the task of "automated detection of quotation errors".

**Methodology -** Adopting a large language model (LLM)-based approach, this paper improves detection performance from two aspects on the basis of existing research: first, employ the fine-tuning approach for LLMs to detect quotation errors; second, incorporating full-text data of the cited literature into dataset construction, and exploring the optimal scheme for building such datasets by comparing three types of full-text integration methods. Based on this, this paper further uses the TokenSHAP tool to conduct interpretability experimental analysis on the model's prediction results.

**Findings -** The fine-tuning approach for LLMs has improved the performance in detecting quotation errors. Among the different methods for incorporating full-text information, the approach based on using the source abstract yielded the best performance.

**Originality -** The fine-tuning approach for large language models (LLMs) is applied to the task of automated detection of quotation errors, and interpretability analysis is conducted on the model's output results.

## 1 Introduction

In the context of accelerating academic exchange and knowledge production, the importance of academic writing standards has become increasingly prominent. Citations, as an integral component of academic papers, serve a dual purpose: they act as a means of tracing knowledge sources(Dontcheva-Navratilova, 2016; Min *et al*., 2021) and substantiating arguments(Alramadan, 2023; Dontcheva-Navratilova, 2016). A citation comprises two fundamental components: annotation information and content information. Annotation information pertains to the citation format, whereas content information refers to the main content of the citation. Citations play an important role in academic research and have been applied in various fields, such as academic evaluation(Waltman, 2016), knowledge diffusion analysis(Yang and Liu, 2022), and interdisciplinary research(Kong *et al*., 2025; Liu and Lu, 2025). The evolution of technologies such as natural language processing has prompted a shift in the focus of citation research, leading to an expansion from citation annotation information to the contextual and content levels of citations. For example, in context of citation-based academic evaluations, the analysis of citation content can be utilized to discern the emotional orientation and intent of citations(Anderson and Lemken, 2023).

As citation content analysis become more widely adopted, issues related to inaccurate citation content have also drawn attention from the academic community. Modifying the original intent of the cited paper can cause inconsistencies between the citation content and the cited paper's original text. Gosling *et al*.(2004) and Neihouse *et al*.(1989) delineate such citation content errors as "quotation errors". Such errors have the potential to lead to a misinterpretation of the original research, resulting in interpretations that deviate from the facts during dissemination. This phenomenon has the potential to influence the academic community's collective understanding of related issues(Greenberg, 2009).

Nowadays, quotation errors remain widespread in academic literature. Cobb *et al*.(2024) examined 3,347 citations from 89 articles in eight prominent psychology journals and found that only 81.2% of the citations were classified as accurate, meaning that nearly 20% of the citations exhibited quotation errors. Smith *et al*.(2020) examined 250 citations in academic journals of high influence such as Nature and Science, finding that the rate of quotation errors reached 25%. This reflects that the phenomenon of quotation errors in high-quality academic publications cannot be ignored.

Existing approaches for the detection of quotation errors can be categorized into three types: similarity-based methods(Liu *et al*., 2024), text classification methods based on small-scale pre-trained language models (PLMs) (Sarol *et al*., 2024), and prompt-based approaches with large language models (LLMs)(Sarol *et al*., 2024). Compared with deep learning methods, the approach of fine-tuning large language models involves more model parameters and exhibits stronger capabilities. Existing studies have shown that, in a subset of natural language processing (NLP) tasks, compared with prompt learning approaches, fine-tuning LLMs exhibits superior performance through task-specific parameter optimization—and also holds advantages in terms of exact match performance and performance improvement for small-scale LLMs(Shang *et al*., 2025; Trad and Chehab, 2024). Therefore, this study introduces the LLM fine-tuning method into the quotation error detection task and adopts LoRA fine-tuning technique(Hu *et al*., 2022) to reduce model training costs. Furthermore, when analyzing the detection results of citation content by the fine-tuned model, this study not only employs conventional quantitative metrics but also conducts manual analysis of the model's misclassification results and proposes optimization pathways. In summary, this study proposes the first research question:

**RQ1**: Can the application of the fine-tuning approach enhance model performance? Is there still potential for further performance improvement of the fine-tuned model?

Second, existing research on quotation error detection primarily verifies the semantic content of target citation sentences using the abstract text of cited paper. Considering that scenarios where only the abstract information of cited paper is provided may lead to insufficient information for quotation error detection, this study additionally constructs a full-text dataset. This dataset includes target citation sentences, abstracts and full-text sentences of cited paper, and corresponding labels. Moreover, this study adopts three types of methods to incorporate full-text information of cited paper, which are screening full-text sentences based on text similarity with abstracts; screening full-text sentences based on text similarity with target citation sentences; and directly incorporating full-text text. These methods are used to compare their impacts on detection performance, providing references for future related experiments. In summary, this study proposes the second research question:

**RQ2**: Can the integration of full-text data from cited paper improve the performance of the fine-tuning approach in detecting quotation errors?

Finally, quotation errors are associated with academic integrity. When quotation errors are detected, the reasons for such judgments should be provided to support subsequent review and correction. Existing research on quotation error detection methods only focuses on numerical evaluation of detection performance, while neglecting the explanation of reasons behind quotation errors. When analyzing the model's detection results of citation content, this study introduces the TokenSHAP method(Horovicz and Goldshmidt, 2024) to conduct interpretability analysis on the output results of LLMs. Unlike traditional SHAP methods(Lundberg and Lee, 2017)—which are primarily used to explain the impact of input features (such as sentence length and sentiment orientation) on model outputs—TokenSHAP is specifically designed for token-level interpretation in text. It quantifies the contribution of each token to the LLM's output using an efficient Monte Carlo Shapley estimation approach. In summary, this study proposes the third research question:

**RQ3**: Can the TokenSHAP method perform interpretability analysis for the model's results?

To conclude, this study makes the following innovations in automated quotation error detection: It applies the method of fine-tuning LLMs to the task of quotation errors detection and conducts manual analysis on the model's misprediction results; It incorporates full-text data from cited paper into this task and employs three full-text integration methods to evaluate the model's performance under different textual granularities; It conducts interpretable experiments and analysis on model's outputs.

## 2 Related Work

This section reviews existing research on quotation error classification, summarizing existing definitions and classification frameworks, and further summarizes key methods for automated quotation error detection while outlining the technical approaches and research developments to date.

### *2.1 Classification of Quotation Errors*

Research on the classification of quotation errors began relatively early, with some studies proposing a binary categorization. For example, Eichorn and Yankauer(1987) in 1987, and Evans *et al*.(1990) in 1990, classified quotation errors into two types: major errors and minor errors. In 2000, Fenton *et al*.(2000) adopted the definition of major errors from the earlier studies, while redefining minor quotation errors as those that exert minimal impact on the meaning of the original source. Subsequent research has proposed increasingly fine-grained classification schemes for quotation errors(Curlewis *et al*., 2022; Lock and Bearman, 2018). However, these classification frameworks often suffer from ambiguous boundaries and overlapping categories in practice, limiting their applicability to automated detection tasks. Wadden *et al*.(2020) categorize the veracity of scientific claims into three classes: SUPPORTS, REFUTES, and NOINFO, deliberately omitting further granularity within the REFUTES category. This simplification treats labels solely as indicators of whether a scientific claim accurately reflects its cited source, thereby enhancing the operational feasibility of quotation error detection.

### 2.2 Detection of Quotation Errors

Existing methods for identifying quotation errors can be broadly categorized into three types: similarity-based methods, deep learning-based methods, and LLM-based approaches.

*(1) Similarity-Based Methods*

Similarity-based methods identify quotation errors by comparing the semantic similarity either between different citation contents or between a citation sentence and the cited paper. Liu *et al*.(2024) employed a BERT-based model to generate sentence embeddings and computed cosine similarity between the citation context and either the full abstract of the cited paper or its segmented sentences. The similarity scores were then used to determine whether quotation errors were present. Liu *et al*.(2021) further applied this approach to detect "indirect citation" quotation errors: in a triangular citation structure where paper C cites both A and B, and B also cites A, they calculated the textual similarity between the citation sentence of B citing A and that of C citing A. If the similarity exceeded 0.9, the method inferred a potential risk of indirect citation.

*(2) Deep Learning-Based Methods*

Currently, research on the automated detection of quotation errors using deep learning methods remains limited. Therefore, this study draws upon scientific claim verification, which is closely related to the detection of quotation errors, as a methodological reference. In 2020, Wadden *et al*.(2020) introduced the task of scientific claim verification and proposed a model consisting of three modules: abstract retrieval, evidence selection, and label prediction. Notably, quotation error detection is highly similar to the tasks of the evidence selection and label prediction modules. Subsequent studies on scientific claim verification have largely focused on optimizing individual sub-modules to enhance overall model performance. Pradeep *et al*.(2021) improved the label prediction module: instead of using the abstract directly as input, they made judgments based on extracted evidence sentences. In addition, joint modeling approaches have been explored to improve overall performance on the scientific claim verification task. Li *et al*.(2021) jointly modeled the evidence selection and label prediction modules, comparing a simple attention mechanism with a kernel graph attention network in the label prediction module. Zhang *et al*.(2021) proposed a joint modeling approach that simultaneously trains abstract retrieval, evidence selection, and label prediction modules to enhance inter-module information flow. During training, the model dynamically adjusts the proportion of gold and predicted labels in the training data to improve performance. Beyond methodological innovations, evaluation perspectives have also evolved. Vladika *et al*.(2024) excluded the "insufficient information" label, which is inconsistently defined across datasets, and conducted binary classification evaluations on multiple datasets using the DeBERTa-v3 model.

*(3) LLM-Based Methods*

LLM-based methods primarily leverage LLMs to detect quotation errors through prompt-based approaches. These approaches are typically categorized into zero-shot and few-shot paradigms. In the zero-shot setting, prompts consist of task instructions, citation content, and the original text of the cited paper. The few-shot setting extends this by including a small number of annotated examples. Existing studies in this domain generally utilize two types of datasets: those containing only abstracts and those containing partial full-text data.

In scenarios where only the abstract of the cited paper is provided, Alvarez *et al*.(2024) transformed the single scientific claim in the SCIFACT dataset into the original citation sentence, retained the two labels Supported and Insufficient Information, and used GPT-3.5 to generate negated citation sentences, thereby constructing the SCitance dataset. They evaluated GPT-3.5 and GPT-4 under both zero-shot and few-shot prompts. Koneru *et al*.(2024) further examined the performance of GPT-3.5 Turbo and PaLM2 by varying task instructions and model temperature parameters.

In scenarios where the full text of the cited paper is introduced, Zhang *et al*.(2024) used a zero-shot prompting approach to evaluate the performance of GPT-3.5 Turbo, GPT-4 Turbo, and GPT-4o under three information conditions: providing (1) the title of the cited paper, (2) the title and abstract, and (3) the title, abstract, and full text. The full-text integration methods were divided into two types: the first segmented the full text into 256 chunks and selected the three chunks most semantically similar to the citation sentence; the second provided the title and used retrieval-augmented generation (RAG)(Lewis *et al*., 2020) via the OpenAI Assistant API. Both supplemented the citation sentence with semantically relevant content from the cited paper and used a prompt-based approach for quotation error detection.

First, although some studies have proposed finer-grained classification schemes for quotation errors, such categories are often difficult to distinguish in practice. As the primary objective is to detect quotation errors rather than assess their severity, this study adopts the classification framework of Wadden et al. Second, prompt-based approaches are adopted, using LLMs for zero-shot or few-shot inference to evaluate their performance in this task. To further enhance LLM effectiveness and adaptability, this study proposes a fine-tuning approach tailored to the detection of quotation errors. Third, most studies focus on quantitative performance metrics, with limited analysis of model outputs and interpretability. In this work, we complement performance comparison with manual analysis and interpretability experiments to improve understanding and trust in model predictions. Finally, in current research on quotation error detection, datasets containing only the abstracts of cited paper are commonly used. Building on this foundation, this study incorporates the full-text information of cited paper into this task, aiming to enhance the performance of quotation error detection by supplementing richer textual semantic context.

# 3 Methodology and Dataset

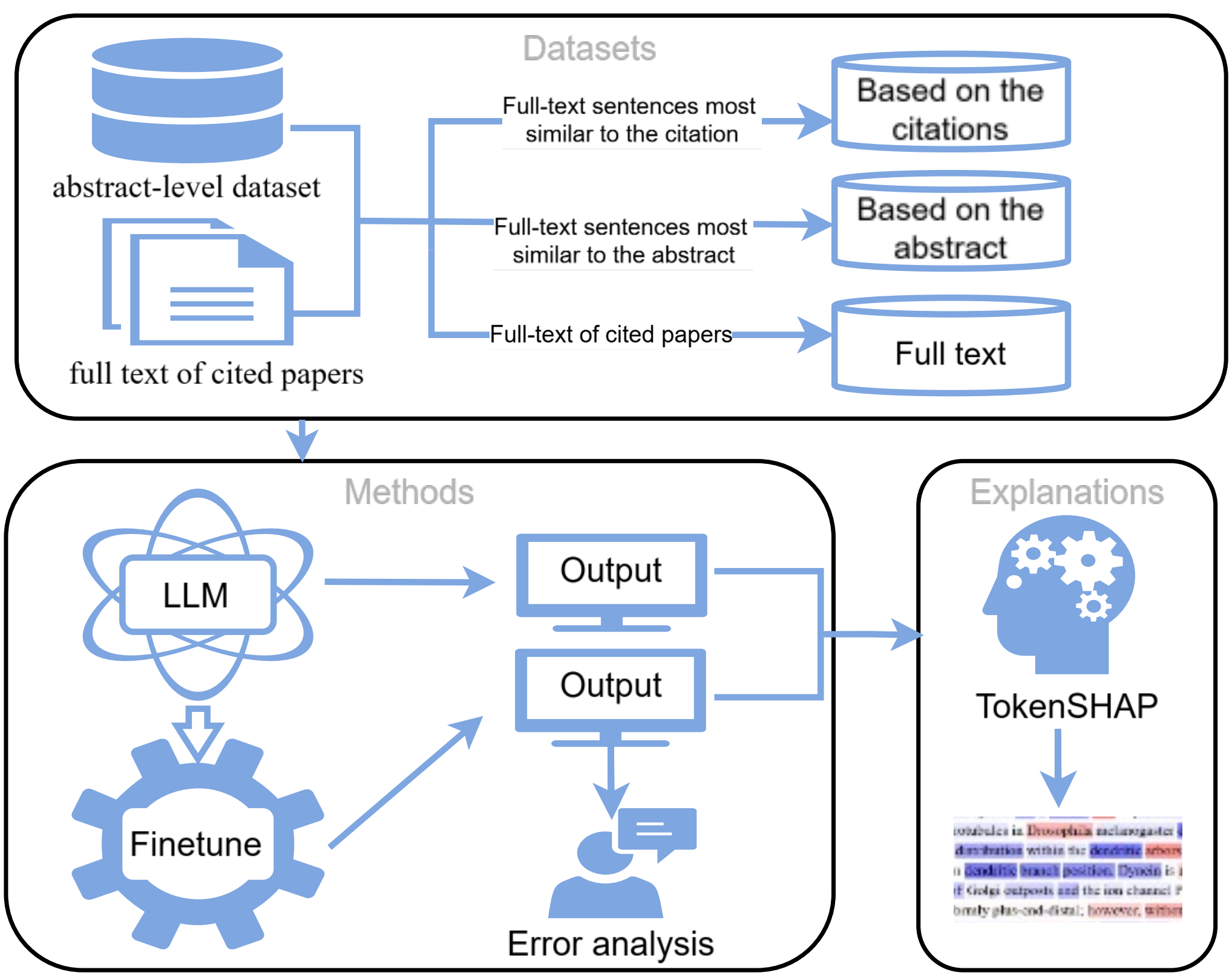

**Figure 1** Research Framework

As illustrated in Figure 1, the research framework proceeds as follows. First, based on the abstract-level dataset, this study constructs a full-text-level dataset using three full-text information integration methods: selecting full-text sentences according to the semantic similarity of citation sentences, selecting full-text sentences according to the abstracts of cited paper, and directly incorporating full-text information. Second, we propose a fine-tuning approach to further improve model performance and conduct error analysis on the incorrect prediction results of the fine-tuned model. Finally, we introduce the TokenSHAP tool to perform interpretability analysis on model outputs, thereby enhancing the interpretability of model outputs.

## *3.1 Fine-Tuning Methods*

We first constructed a standardized training data format tailored to the detection of quotations errors, which integrates four core components: task instructions, the target citation sentence $C_i$, the cited paper information $R_i$ and the corresponding label $L_i$.

Next, to balance the performance improvement of the LLM on the specific task and the efficiency of model training, we applied LoRA fine-tuning(Hu *et al.*, 2022) to the LLM, configuring both the LoRA parameters and general training parameters. These parameter settings were determined through preliminary experimental validation to ensure that the model can converge stably during training and avoid problems such as overfitting or slow convergence. The model was then trained and checkpointed during the training process.

Finally, to comprehensively evaluate the performance of the fine-tuned model on the quotation error detection task, we constructed a dedicated test dataset that maintains consistency with the training data in terms of structure, which consists of task instruction, target citation sentence $C_i$, and cited paper information $R_i$, which produced prediction outputs accordingly.

Additionally, to verify the performance improvement brought by fine-tuning, we tested the pre-trained model using the same dataset and compared it with the fine-tuned model.

### *3.2 Data*

#### *(1) Selection of Raw Datasets*

We selected two open-source datasets for manual annotation: the SCIFACT released by Wadden *et al*.(2020), and the NealSmith-2024 released by Smith *et al*.(2020).

**SCIFACT.** Each sample comprises a citation sentence (decomposed into a single scientific claim), the cited paper abstract, and a label. A scientific claim is defined as the minimal, indivisible, and verifiable unit of scientific content. Labels are categorized into three classes: SUPPORTS, REFUTES, and NOINFO.

**NealSmith.** This dataset contains 250 samples. Each sample includes metadata from the citing paper, the title and DOI of the cited paper, annotator information, and the annotation label. Samples are categorized into four classes: Fully Substantiated, where the citation is fully supported by the cited paper; Partially Substantiated, indicating minor inaccuracies; Unsubstantiated, where the citation contradicts the cited content; and Impossible to Substantiate, where verification is not feasible based on the cited paper.

#### *(2) Dataset Annotation Process*

First, unlabeled samples and those with unavailable full text are removed from the original dataset to build the study's abstract-level dataset. For the SCIFACT dataset, this study screens 1,182 samples from the training and development sets. For the NealSmith dataset, 222 valid samples are selected. For citation sentences with multiple references, remove content related to other references; the rest stays unchanged. In this phase, the study's abstract-level datasets SCIFACT-AB and NealSmith-AB are constructed.

The second, the top 10 sentences with the highest semantic similarity to the abstract are extracted from the full text of the cited paper. These sentences are appended to the abstract of the cited source to construct the full-text datasets: SCIFACT-FULL and NealSmith-FULL.

Finally, the labels of the SCIFACT dataset and the NealSmith dataset are standardized. For the SCIFACT dataset, we retain the original three-class annotation scheme, renaming the labels SUPPORTS, REFUTES, and NOINFO as SUPPORT, CONTRADICT, and NULL. For the NealSmith dataset, since the quotation error detection task focuses on identifying errors rather than their severity, both "Partially Substantiated" and "Unsubstantiated" categories indicate inconsistencies between the citation content and the cited paper, with the only difference being the degree of inconsistency. Therefore, the labels Partially Substantiated and Unsubstantiated are merged into a single category, indicating contradiction between the citation content and the cited paper. Accordingly, the dataset is re-annotated into three classes: Fully Substantiated is mapped to SUPPORT, Partially Substantiated and Unsubstantiated to CONTRADICT, and Impossible to Substantiate to NULL.

# 4 Results Analysis

## 4.1 Baseline for Automated Detection of Quotation Errors

### (1) Based on PLMs

**SCIBERT model.** In the data input module, the target citation sentence $C_i$, and the abstract or full-text data of the cited paper as $R_i$ are first input into the SCIBERT model, and after being processed by the model, the predicted classification label is finally output.

**T5 model**(Raffel *et al.*, 2020). In the data input module, the target citation sentence $C_i$ and the abstract or full-text data of the cited paper $R_i$ and the prompt $P$ are concatenated and fed into the T5 model. The T5 model processes the input and generates the corresponding classification label.

### (2) Prompt-Based

System-level information includes task instructions, while the user-level information includes the target citation sentence $C_i$ and the information from the cited paper $R_i$(Alvarez *et al.*, 2024) .

In the prompt-based experimental setup, we adopt three paradigms: zero-shot, few-shot (3-shot)(Brown et al., 2020), and retrieval-augmented generation (RAG), respectively. Among them, the 3-shot setting provides the model with demonstration examples corresponding to three types of labels. Under the RAG framework, two retrieval-augmented strategies are employed in this study: (i) Dense-sparse hybrid retrieval based on Reciprocal Rank Fusion (RRF)(Cormack et al., 2009): BM25 for sparse retrieval and PubMedBERT for semantic dense retrieval are separately used to recall relevant passages from full-text papers, and the RRF algorithm is adopted to fuse and re-rank the multi-source results. This method is abbreviated as "hybrid" in Table III. (ii) Hypothesis-driven document embedding retrieval(Gao et al., 2023): a generative language model is utilized to construct hypothetical evidence related to the claim to be verified. Three separate semantic retrievals are performed with the original claim, supportive hypothesis, and refutative hypothesis, and the Top-6 relevant passages are finally returned. This method is abbreviated as "hyde" in Table III.

## 4.2 Model Settings

### (1) Model Selection

For the selection of PLMS, we adopted a SCIBERT model and the T5 model. Prior studies have demonstrated that BERT-based models achieve promising performance on scientific claim verification tasks (Wadden *et al.*, 2020). Given that all texts involved in this study are derived from scientific literature, we selected SCIBERT—an extended model built on the BERT architecture and pre-trained on large-scale scientific corpora—for its domain-specific optimization. As for the T5 model, which features an encoder–decoder architecture, its unifiedqa-large version was selected for our study.

For prompt learning in this study, the models Qwen-Plus (QPlus), Qwen-Max (QMax), DeepSeek-V3 (DeepSeek-AI, 2024b), and DeepSeek-R1(DeepSeek-AI, 2024a) are used. Qwen-Plus adopts a decoder-only architecture and is suitable for tasks of medium complexity such as text generation; Qwen-Max is suitable for complex multi-step tasks such as deep reasoning; DeepSeek-V3(DV3) is a mixture-of-experts LLM; DeepSeek-R1(DR1) is a model series focused on enhancing reasoning capability.

This study fine-tunes Qwen-2.5-7B (Q7B) and DeepSeek-R1-Distill-Llama-8B (D-L8B) respectively. Qwen-2.5-7B is an LLM with a decoder-only architecture; DeepSeek-R1-Distill-Llama-8B is distilled from the DeepSeek-R1 model based on the Llama architecture, further optimizing the reasoning and task-handling capabilities of the original Llama-8B model.

*(2) Experimental Parameter Settings*

For the SCIBERT-based model, the number of training epochs is set to 20, with a batch size of 32 and a learning rate of 3e-5. For the T5 model, the number of training epochs is set to 10, with a batch size of 1 and a learning rate of 1.5e-4.

For Prompt-based approach, the Qwen-Plus model version used is qwen-plus-2025-04-28, and the Qwen-Max model version is qwen-max-2025-01-25. The DeepSeek series models were accessed on June 25, 2025.

For the fine-tuning approach, we employ LoRA-based fine-tuning for large language models. On the SCIFACT dataset, models are trained for 10 epochs with a batch size of 1, a learning rate of 1e-4, and a gradient accumulation step of 4. To mitigate overfitting on the smaller NealSmith dataset, we use fewer training epochs: 8 epochs for the D-L8B model and 3 epochs for the Q7B model.

*(3) Model Performance Evaluation and Analysis Methods*

The evaluation of model performance adopts Precision (P), recall (R), F1 score, and macro-average. Macro-average is defined as the arithmetic mean of each metric across all labels.

To comprehensively evaluate LLM performance in quotation error detection, we conduct error analysis on misclassified predictions of fine-tuned models, explore the root causes of such errors in depth, and propose optimization paths for model improvement.

For the model interpretability analysis, the TokenSHAP method is employed in this study. The TokenSHAP method estimates the marginal contribution of each token in the input prompt to the model output, thereby quantitatively assessing its importance to the generated result. First, the model is required to output brief justifications in the prompt. Then, individual tokens in the prompt are removed one at a time, and the modified prompts are input into the LLM to obtain corresponding outputs. Second, the cosine similarity between the TF-IDF vector of each modified output and the output obtained from the complete prompt is computed. The average cosine similarity for outputs containing a given token is subtracted from the cosine similarity for outputs without that token to yield its contribution value. Finally, the token contributions are visualized.

## *4.3 Experimental Results*

*(1) Evaluation Results of Detection Performance*

First, regarding the performance of SCIBERT and T5 (Table I), a consistent pattern emerges across both abstract-level and full-text-level datasets: the T5 model achieves better performance on the SCIFACT dataset, while the SCIBERT-based model performs better on the NealSmith dataset. Specifically, T5 achieves an 80.20% macro-averaged F1-score on SCIFACT's abstract-level dataset, outperforming SCIBERT by 6.55%. Conversely, on the NealSmith abstract-level dataset, the SCIBERT-based model achieved a macro-averaged F1 score of 50.37%, surpassing T5 by 2.26%. On the SCIFACT full-text dataset, the T5 achieved a macro-averaged F1 score of 78.32%, exceeding SCIBERT by 5.67%. On the NealSmith full-text dataset, the SCIBERT-based model reached a macro-averaged F1 score of 53.20%, outperforming T5 by 5.83%. At the category level, T5 has a clear advantage over SCIBERT in predicting the "CONTRADICT" class on SCIFACT, while SCIBERT outperforms T5 in this class on NealSmith.

**Table I** Results (%) of PLMs

| Dataset | Model | SUP. | CON. | NUL. | Macro-Average |
|---|---|---|---|---|---|
| | | $F_1$ | $F_1$ | $F_1$ | $F_1$ |
| SCIFACT-AB | SCIBERT | 72.46 | 59.28 | **89.20** | 73.65 |
| | T5 | **83.60** | **73.16** | 83.85 | **80.2** |
| NealSmith-AB | SCIBERT | 81.98 | **21.79** | 47.31 | **50.37** |
| | T5 | **83.89** | 4.44 | **56.00** | 48.11 |
| SCIFACT-FULL | SCIBERT | 71.64 | 58.93 | **87.39** | 72.65 |
| | T5 | **80.41** | **71.82** | 82.72 | **78.32** |
| NealSmith-FULL | SCIBERT | 79.67 | **29.41** | 50.51 | **53.20** |
| | T5 | **85.67** | 5.00 | **53.43** | 47.37 |

Second, regarding the performance of prompt-based approaches and fine-tuning approaches using LLMs, Tables II and III show the experimental data on the abstract-level dataset and the full-text-level dataset respectively. Entries marked with "*" denote statistically significant differences. "FT" indicates that the method adopted is Fine-Tuning, "(3-shot)" represents 3-shot prompt learning. In addition, the model used for retrieval-augmented generation (RAG) is Qwen-plus.

**In the SCIFACT dataset, the fine-tuning approach based on the Qwen-7B model demonstrates superior performance.** Its macro-averaged F1 scores on the abstract-level dataset and the full-text-level dataset are 87.37% and 86.54%, respectively. These scores exceed those of the best-performing model among the prompt-based approaches—Qwen-Max—by 2.43% (abstract-level dataset) and those of the best-performing model in the full-text-level dataset—DeepSeek-V3—by 4.18%. **In the NealSmith dataset, the fine-tuning approaches based on the DeepSeek-R1-Distill-Llama-8B model and the Qwen-7B model achieve the best performance on the abstract-level and full-text-level datasets, respectively,** with macro-averaged F1 scores of 48.06% and 50.66%, surpassing the best-performing prompt-based approach by 11.41% and 7.34%, respectively.

**Table II** Results (%) of LLM-based methods on the abstract-level dataset

| Dataset | Model | SUP. | CON. | NUL. | Macro-Average |
|---|---|---|---|---|---|
| | | $F_1$ | $F_1$ | $F_1$ | $F_1$ |
| SCIFACT-AB | DV3 | 89.17 | 84.71 | 79.25 | 84.38 |
| | DR1 | 87.14 | 82.22 | 72.81 | 80.72 |
| | QPlus | 82.84 | 79.66 | 63.97 | 75.49 |
| | QMax | **89.55** | **85.62** | 79.65 | **84.94** |
| | DV3(3-shot) | 85.33 | 80.05 | 78.78 | 81.39 |
| | DR1(3-shot) | 65.88 | 39.37 | 32.63 | 45.96 |
| | QPlus(3-shot) | 85.92 | 79.04 | 72.51 | 79.16 |
| | QMax(3-shot) | 87.57 | 82.42 | **80.19** | 83.39 |
| | **Q7B** | 63.24 | 9.08 | 34.55 | 35.62 |
| | **D-L8B** | 77.56 | 73.83 | 56.82 | 69.40 |
| | **Q7B(FT)** | **89.90** | **82.50** | **90.61** | **87.37***|
| | **D-L8B(FT)** | 78.81 | 67.32 | 84.23 | 76.79 |
| NealSmith-AB | DV3 | 60.99 | 13.00 | 22.93 | 33.30 |
| | DR1 | 69.03 | **16.97** | 23.95 | **36.65** |

| | | | | |
|---|---|---|---|---|
| QPlus | 76.89 | 15.64 | 15.22 | 35.91 |
| QMax | 65.71 | 12.44 | 25.13 | 34.43 |
| DV3(3-shot) | 63.59 | 8.89 | 27.10 | 33.19 |
| DR1(3-shot) | 82.83 | 4.44 | 0 | 29.09 |
| QPlus(3-shot) | 72.97 | 14.55 | 18.26 | 35.26 |
| QMax(3-shot) | 63.73 | 05.00 | 26.81 | 31.85 |
| **Q7B** | 77.06 | 0 | 25.09 | 34.05 |
| **D-L8B** | 76.71 | 4.44 | 15.85 | 32.34 |
| **Q7B(FT)** | **82.49** | 18.76 | **41.24** | 47.49* |
| **D-L8B(FT)** | 81.92 | **23.24** | 38.03 | **48.06*** |

When comparing various prompt-based approaches, it is found that the introduction of in-context learning and Retrieval-Augmented Generation (RAG) does not significantly improve model performance. In contrast, the difference in the macro-averaged F1 scores of the model before and after fine-tuning intuitively reflects the significant improvement effect of the fine-tuning strategy on model performance. Examination of the F1 scores by category reveals that the fine-tuned Qwen-7B model achieves consistently higher performance across the “CONTRADICT” and “NULL” categories compared to other models. Additionally, the fine-tuned DeepSeek-R1-Distill-Llama-8B model also achieves higher macro-averaged F1 scores than the prompt-based approaches. This result indicates that the method of fine-tuning LLM can improve the performance of the detection of quotation errors, address the first research question.

**Table III** Results (%) of LLM-based methods on the full-level dataset

| Dataset | Model | SUP. | CON. | NUL. | Macro-Average |
|---|---|---|---|---|---|
| | | $F_1$ | $F_1$ | $F_1$ | $F_1$ |
| SCIFACT-FULL | DV3 | **87.6** | **83.95** | 75.53 | **82.36** |
| | DR1 | 83.91 | 80.10 | 69.04 | 77.68 |
| | QPlus | 82.19 | 77.29 | 53.28 | 70.92 |
| | QMax | 87.11 | 81.32 | 73.39 | 80.61 |
| | DV3(3) | 85.56 | 82.33 | 77.33 | 81.74 |
| | DR1(3) | 64.93 | 40.58 | 20.47 | 42.00 |
| | QPlus(3) | 82.53 | 74.41 | 58.73 | 71.89 |
| | QMax(3) | 86.56 | 80.31 | 77.00 | 81.29 |
| | Rag(hybird) | 81.44 | 75.89 | 47.57 | 68.3 |
| | Rag(hyde) | 79.07 | 73.22 | 43.81 | 65.37 |
| | **Q7B** | 60.98 | 2.81 | 14.60 | 26.13 |
| | **D-L8B** | 77.35 | 71.87 | 56.20 | 69.40 |
| | **Q7B(FT)** | **88.19** | **83.00** | **88.79** | **86.54*** |
| | **D-L8(FT)** | 63.02 | 49.80 | 57.29 | 56.70 |
| NealSmith-FULL | DV3 | 68.61 | 11.27 | 23.15 | 34.34 |
| | DR1 | 69.36 | 14.35 | 20.55 | 34.75 |
| | QPlus | 80.08 | **19.33** | 14.83 | 38.08 |
| | QMax | 71.14 | 10.30 | 27.18 | 36.20 |
| | DV3(3) | 73.37 | 8.89 | **32.47** | 38.24 |

| | | | | |
|---|---|---|---|---|
| DR1(3) | **83.88** | 0 | 9.00 | 30.96 |
| QPlus(3) | 79.59 | 13.33 | 14.92 | 35.95 |
| QMax(3) | 69.46 | 5.71 | 25.47 | 33.55 |
| Rag(hybird) | 77.51 | 16.65 | 31.23 | 41.80 |
| Rag(hyde) | 76.92 | 24.28 | 28.74 | **43.32** |
| **Q7B** | 82.80 | 0 | 30.83 | 37.88 |
| **D-L8B** | 82.46 | 10.71 | 27.08 | 40.08 |
| **Q7B(FT)** | **85.67** | **23.74** | **42.56** | **50.66*** |
| **D-L8B(FT)** | 82.88 | 14.00 | 37.80 | 44.89 |

To better illustrate the performance improvement brought by fine-tuning, this paper takes the fine-tuning of Qwen-7B on the SCIFACT-AB dataset as an example, and analyzes the changes in model prediction results on the same set of samples before and after fine-tuning. The original Qwen model without fine-tuning predicted 87% of samples as “SUPPORT” and failed to distinguish the “NULL” and “CONTRADICT” classes. For instance, given the claim “90% of SIDS deaths happen in newborns aged less than 6 months”, the corresponding abstract contains no mention of “90%” or “6 months”; yet the model predicted “SUPPORT” merely because the topic of SIDS was present. Similarly, for the claim “A deficiency of vitamin B12 decreases blood levels of homocysteine”, the abstract contains no reference to vitamin B12 whatsoever, but the model still output “SUPPORT”. In contrast, the fine-tuned model correctly classified all such cases. For “CONTRADICT” samples, such as the claim “Autophagy deficiency in the liver increases vulnerability to insulin resistance”, the abstract presents evidence directly contradicting the claim. However, the non-fine-tuned Qwen model still predicted “SUPPORT” based on shallow keyword overlap of “Autophagy deficiency” and “insulin” between the claim and the abstract.

By comparing label distributions before and after fine-tuning, we observe that the pre-fine-tuned model relies on shallow keyword matching and tends to predict “SUPPORT” whenever topical overlap exists. In comparison, the fine-tuned model effectively evaluates the sufficiency of evidence and the consistency of semantic orientation, leading to more accurate citation error detection.

*(2) Error Analysis and Optimization Paths*

The first category is inconsistent prior knowledge between the model and annotators. When the cited paper only provides the infection rate in the UK, annotators consider the UK population size as a known condition and thus label the result as “CONTRADICT”. However, the model determines the result as “NULL” because the paper does not explicitly state the UK population size. The second category is limitations of the model in handling quantitative relationships in text. For the expression “75 nmol/liter” in the citation sentence, if the unit of measurement of this value changes in the abstract of the cited literature, it will directly affect the accuracy of the model's judgment. The third category is insufficient understanding of the concepts of “increment” and “absolute quantity” by the model. The target claim clearly states that “Nigerian physicians constitute the largest group of physicians trained in sub-Saharan Africa within the United States”, where the core lies in the comparison of “absolute quantity”. Nevertheless, the abstract of the cited literature only mentions that the “increment” of Nigerian physicians is the largest, leading to misjudgment by the model. The fourth category is incomplete understanding of sentence semantics by the model. During the recognition process, the model may ignore some key conditions in the citation sentence. For example, in Sample 4 of Table IV, the model omitted the key condition of “BMP4, activin A, CHIR99021, and SU504” during recognition.

**Table IV** Example of LLM prediction errors

| Sample ID | Citation Sentence | Ground Truth | Model Prediction |
|---|---|---|---|
| 1 | A total of 1,000 people in the UK are asymptomatic carriers of vCJD infection. | CON. | NUL. |
| 2 | High dietary calcium intakes are unnecessary for prevention of secondary hyperparathyroidism in subjects with 25(OH)D levels above 75 nmol/liter. | NUL. | SUP. |
| 3 | Nigerian physicians constitue the largest component of sub-Saharan Africa-trained physicians in the United States | NUL. | SUP. |
| 4 | Addition of BMP4, activin A, CHIR99021, and SU504 to reprogramming fibroblasts generates, expands and maintains cardiovascular progenitor cells (CPCs). | SUP. | NUL. |

Based on the above-identified causes of prediction errors in LLMs, we propose the following optimization paths: First, to address the insufficient prior knowledge of LLMs, methods such as retrieval-augmented generation (RAG), tool invocation, and API calls can be employed. Second, to mitigate the limitations of LLMs in recognizing and reasoning about numerical relationships, numerical reasoning pretraining or instruction tuning can be introduced. These strategies assist LLMs in better identifying quantitative relationships within text. Third, regarding the confusion exhibited by LLMs between core semantic concepts, training can be conducted using word pairs that are semantically similar but logically distinct. Finally, for errors arising from incomplete semantic comprehension or partial information alignment, chain-of-thought (CoT) prompting can be integrated.

*(3) Comparison of Full-Text Integration Methods*

To further analyze how different full-text integration methods affect experimental results, we extend the NealSmith full-text dataset (built on similarity to the cited paper’s abstract, referred to as Neal-M1 in this section) by adding two more datasets: Neal-M2 (top 10 full-text sentences with highest textual similarity to the target citation sentence) and Neal-M3 (direct incorporation of the cited paper’s complete full text). Due to the excessively large input length of the Neal-M3 dataset, we evaluated it only using the Qwen-Plus-2025-04-28 model, which supports ultra-long context inputs, and excluded it from evaluation under the fine-tuning approach. In addition, since the SCIFACT dataset is annotated based on abstract text, the introduction of full-text information does not lead to performance improvement; thus, only the NealSmith dataset is adopted in this experiment.

The experimental results on these three datasets are shown in Table V. In prompt-based approaches, Neal-M2 achieved the highest macro-averaged F1 score of 38.14%, which is only 0.06% higher than Neal-M1. However, in fine-tuning approaches, Neal-M1 significantly outperformed Neal-M2, achieving a macro-averaged F1 score of 50.66%. In prompt-based approaches, Neal-M3 yielded a macro-averaged F1 score of only 35.06%, lower than that of the other two full-text-level datasets. Combined with the aforementioned experiments, this study also addresses the second research question—incorporating full-text information can enhance the performance of quotation error detection, and different approaches to full-text integration exert an impact on task performance.

**Table V** Results of Different Cited paper Information Integration in NealSmith (%)

| Dataset | Model | SUP. | CON. | NUL. | Macro-Average |
|---|---|---|---|---|---|
| | | $F_1$ | $F_1$ | $F_1$ | $F_1$ |
| Neal-M1 | Qplus20250425 | 80.08 | 19.33 | **14.83** | 38.08 |
| Neal-M2 | | 78.94 | **19.58** | 13.79 | **38.14** |
| Neal-M3 | | **82.59** | 10.91 | 11.67 | 35.06 |
| Neal-M1 | **Q7B(FT)** | **85.67** | 23.74 | **42.56** | **50.66** |
| Neal-M2 | | 84.31 | **24.71** | 26.43 | 45.15 |
| Neal-M1 | **D-L8B(FT)** | **82.88** | 14.00 | **37.80** | **44.89** |
| Neal-M2 | | 82.12 | **17.46** | 27.14 | 42.24 |

*(4) Overfitting Verification of the Trained Model*

**Table VI** Results of the fine-tuned model on the training and test sets (%)

| Dataset | Model | SUP. | CON. | NUL. | Macro-Average |
|---|---|---|---|---|---|
| | | $F_1$ | $F_1$ | $F_1$ | $F_1$ |
| Train | **Q7B(FT)** | 90.26 | 18.25 | 48.74 | 52.43 |
| Test | | 82.49 | 18.76 | 41.24 | 47.49 |
| Train | **D-L8B(FT)** | 86.64 | 43.13 | 32.93 | 54.23 |
| Test | | 81.92 | 23.24 | 38.03 | 48.06 |

Due to the limited size of the NealSmith dataset, to verify whether the fine-tuned model suffers from overfitting, this section presents a performance comparison of NealSmith-AB on two fine-tuned models, with results shown in Table VI.

It can be observed from the table that although the F1-score on the training set is slightly higher than that on the test set, the difference lies within a reasonable range, indicating that no obvious overfitting occurs in the model.

### *4.4 Interpretability Analysis*

In this section, the TokenSHAP(Horovicz and Goldshmidt, 2024) method is employed to visualize the outputs of the prompt-based approach, with representative cases shown in Figure 2. Figure 2-(a) presents an instance where the prompt learning model generates an incorrect output, misclassifying "CONTRADICT" as "NULL". In contrast, Figure 2-(b) demonstrates the correct output of the fine-tuned model for the same instance, with the model outputting "CONTRADICT".

Figure 2 illustrate the contribution of individual tokens from the target citation sentence and the abstract to the model's final output. Terms highlighted in red indicate that the vocabulary has a promoting effect on the model's output label, while terms highlighted in blue tend to guide the model to output other labels. The darker the color, the stronger the degree of influence.

In the citation sentence of Figure 2-(a), terms such as "Upregulation", "Drosophila", "increases", and "number" are highlighted in red, while "microtubule" is highlighted in blue. This indicates that the content of the citation sentence focused on by the model is "the upregulation of a certain substance in Drosophila increases the number of microtubule plus-ends growing toward the neuron cell body", and the model pays more attention to the content related to "increase" in the citation sentence. In the abstract section, although terms indicating importance (e.g., "role" and "necessary") appear, this information does not directly reveal content relevant to the citation sentence. Additionally, some terms that seem relevant to the citation sentence on the surface (e.g., "Drosophila" and "minus-end") fail to support the causal relationship proposed in the citation sentence under the context of the abstract. This makes the model more inclined to determine that the abstract information is insufficient to support the citation sentence. In particular, the term "without" preceding "dynein" further guides the model to output "NULL".

The TokenSHAP visualization output of the fine-tuned model for the same sample is shown in Figure 2-(b). The output category of this sample is "CONTRADICT"; therefore, the analysis focuses on the red-highlighted terms that make a positive contribution to the model's conclusion of "CONTRADICT". In the citation sentence, the "plus-ends" and "toward" are highlighted in red, indicating that the model accurately captures the core detail of the claim—the orientation feature of microtubule plus-ends—and this feature is precisely the key to the semantic inconsistency between the claim and the abstract of the cited paper. In the abstract section, the sentence containing "uniformly" clarifies the basic fact that "axonal microtubules maintain a plus-end-distal orientation under normal conditions", while the two sentences containing "without" (indicating absence) further point out that "axonal microtubules exhibit polarity disorganization when dynein is absent". The above content in the abstract forms an explicit semantic opposition to the statement in the claim that "upregulation of dynein increases the number of microtubule plus-ends growing toward the cell body". The fine-tuned model successfully captures this core semantic difference and thus classifies the sample as "CONTRADICT".

The results address the third research question: the TokenSHAP method enables interpretable analysis of the models. The visualization results of TokenSHAP indicate that, the fine-tuned model can not only identify the key semantic information in the abstract that is relevant to the citation sentence, but also further conduct reasoning and analysis on the semantic association between the abstract and the citation sentence, thereby improving the accuracy of the quotation error detection task.

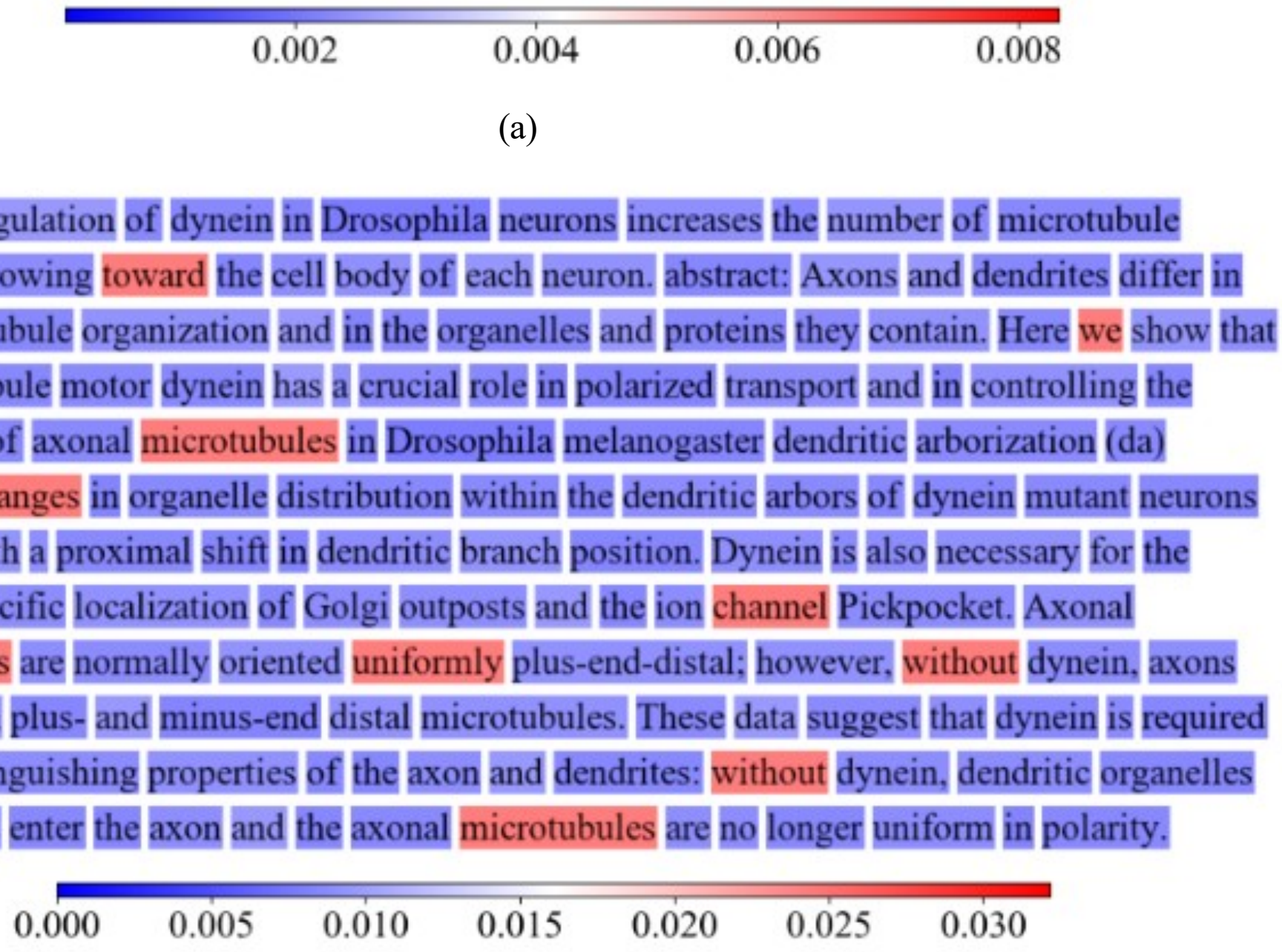

**Figure 2** TokenSHAP Visualization Output

## 5 Conclusion and Future Work

This study aims to evaluate the performance of LLM-based methods in the automated detection of quotation errors in academic literature, as well as to explain and analyze the output results of LLMs. Centered on this objective, three research questions are proposed. The specific research findings and subsequent analysis are as follows:

Regarding the first research question, the fine-tuning approach for LLMs outperforms other baseline models across different datasets. Furthermore, this study performs manual analysis on the misprediction results of LLMs, roughly classifying them into four error types and proposing corresponding optimization paths.

Regarding the second research question, incorporating full-text data can enhance task performance, and among the various methods for integrating the full text of cited papers, the approach of "selecting full-text sentences based on textual similarity with abstracts" achieves the optimal performance. Notably, this conclusion is drawn solely from the results on the NealSmith dataset. Since the SCIFACT dataset is annotated based on abstracts only, it is not possible to achieve higher accuracy by introducing full-text information. In contrast, the NealSmith dataset is annotated with full-text context, which is more consistent with the actual scenario of detection of quotation errors.

For the third research question, the TokenSHAP method can help gain a preliminary understanding of the model's output results.

This study still has several limitations. First, when reorganizing the NealSmith dataset into NealSmith_AB (abstract-level) and NealSmith_FULL (full-text-level), two of the original four classification labels were directly merged into one category to construct a three-way classification dataset. The imprecision of this label conversion may have affected the experimental performance. Second, the sample size of the NealSmith dataset used is relatively small; future research should incorporate larger-scale citation datasets to support more convincing experiments and validations. Finally, the current work on LLM fine-tuning remains in the preliminary exploration stage, leaving ample room for in-depth research.

In future work, we will improve task performance through the following methods: expanding the dataset via data augmentation and manual annotation, introducing chain-of-thought to guide the model in reasoning about and understanding the semantic information of text, and incorporating noise-enhanced training during model training.

## Acknowledgement

This paper was supported by the National Social Science Fund of China (Grant No.24FYB077).

## Data availability

The data sample and source code used in this study are publicly available in a GitHub repository: https://github.com/hlurry/Automated-Detection-of-quotation-errors.

## Declarations

Conflict of interest: The author(s) declared no potential conflicts of interest with respect to the research, authorship, and/or publication of this article.

## References:

Alramadan, M. M. (2023). "Citation behavior, audience awareness, and identity construction in Arabic and EFL research", *Heliyon*, Vol.9 No.2, e13125, doi: 10.1016/j.heliyon.2023.e13125

Alvarez, C., Maxwell Bennett, and Wang, L. (2024), "Zero-shot Scientific Claim Verification Using LLMs and Citation Text", in *Proceedings of the Fourth Workshop on Scholarly Document Processing* (SDP 2024), Bangkok, Thailand: Association for Computational Linguistics, pp. 269–276, available at: https://aclanthology.org/2024.sdp-1.25/ (accessed 20 June 2025)

Anderson, M. H., and Lemken, R. K. (2023), "Citation Context Analysis as a Method for Conducting Rigorous and Impactful Literature Reviews", *Organizational Research Methods* , Vol. 26, No. 1, pp. 77–106, doi: 10.1177/1094428120969905

Brown, T. B., Mann, B., Ryder, N., Subbiah, M., Kaplan, J., Dhariwal, P., Neelakantan, A., Shyam, P., Sastry, G., Askell, A., Agarwal, S., Herbert-Voss, A., Krueger, G., Henighan, T., Child, R., Ramesh, A., Ziegler, D. M., Wu, J., Winter, C., Hesse, C., Chen, M., Sigler, E., Litwin, M., Gray, S., Chess, B., Clark, J., Berner, C., McCandlish, S., Radford, A., Sutskever, I., and Amodei, D. (2020). "Language models are few-shot learners", in Larochelle, H., Ranzato, M., Hadsell, R., Balcan, M. F. and Lin, H. (Eds), *Advances in Neural Information Processing Systems 33* (NeurIPS 2020), Curran Associates, Inc., Red Hook, NY, USA , pp. 1877–1901, doi: 10.5555/3495724.3495883

Cobb, C. L., Crumly, B., Montero-Zamora, P., Schwartz, S. J., and Junior, C. R. M. (2024), "The Problem of Miscitation in Psychological Science: righting the Ship." *American Psychologist*, Vol. 79, No. 2, pp. 299–311, doi: 10.1037/amp0001138

Cormack, G. V., Clarke, C. L. A., and Buettcher, S. (2009), "Reciprocal rank fusion outperforms condorcet and individual rank learning methods", in *Proceedings of the 32nd International ACM SIGIR Conference on Research and Development in Information Retrieval (SIGIR '09)* , Association for Computing Machinery, Boston, Massachusetts, USA, pp. 758–759, doi: 10.1145/1571941.157211

Curlewis, K., Leung, B., Sinclair, L., Ricketts, D., and Rogers, B. (2022). "Quotation errors related to the wound management of open lower limb fractures (WOLLF) randomized clinical trial", *European Journal of Orthopaedic Surgery and Traumatology*, Vol. 33, No. 4, pp. 701–707, doi: 10.1007/s00590-022-03243-w

DeepSeek-AI. (2024a). "DeepSeek-R1 Incentivizing Reasoning Capability in LLMs via Reinforcement Learning", *arXiv Preprint, arXiv:*2501.12948, doi: 10.48550/arXiv.2501.12948

DeepSeek-AI. (2024b). "DeepSeek-V3 Technical Report", *arXiv Preprint arXiv:*2412.19437, doi: 10.48550/arXiv.2412.19437

Dontcheva-Navratilova, O. (2016), "Rhetorical functions of citations in linguistics research articles: a contrastive (English-Czech) study", *Discourse and Interaction*, Vol. 9 No. 2, pp. 51-74, doi: 10.5817/DI2016-2-51

Eichorn, P., and Yankauer, A. (1987), "Do authors check their references? A survey of accuracy of references in three public health journals", *American Journal of Public Health*, Vol. 77 No. 8, pp. 1011–1012, doi: 10.2105/ajph.77.8.1011

Evans, J. T., and Burchell, A. (1990), "Quotational and reference accuracy in surgical journals: a continuing peer review problem", *JAMA*, Vol. 263 No. 10, pp. 1353–1354, doi: 10.1001/jama.1990.03440100059009

Fenton, J. E., Brazier, H., Souza, A., Hughes, J. P., and McShane, D. P. (2000). "The accuracy of citation and quotation in otolaryngology/head and neck surgery journals", *Clinical Otolaryngology and Allied Sciences*, Vol. 25 No. 1, pp. 40–44, doi: 10.1046/j.1365-2273.2000.00322.x

Gao, L., Ma, X., Lin, J., and Callan, J. (2023), "Precise zero-shot dense retrieval without relevance labels", *Proceedings of the 61st Annual Meeting of the Association for Computational Linguistics (Volume 1: Long Papers)*, Vol. 1, pp. 1762–1777, Association for Computational Linguistics, Toronto, Canada, doi: 10.18653/v1/2023.acl-long.99.

Gosling, C. M., Cameron, M., Gibbons, P. F., and *et al*. (2004), "Referencing and quotation accuracy in four manual therapy journals", *Manual Therapy*, Vol. 9 No. 1, pp. 36–40, doi: 10.1016/s1356-689x(03)00056-0

Greenberg, S. A. (2009), "How citation distortions create unfounded authority: Analysis of a citation network", *BMJ*, Vol. 339, Jul 20, b2680, doi: 10.1136/bmj.b2680

Horovicz, M., and Goldshmidt, R. (2024), "TokenSHAP: Interpreting Large Language Models with Monte Carlo Shapley Value Estimation", *Proceedings of the 1st Workshop on NLP for Science (NLP4Science)*, pp. 1–8, Association for Computational Linguistics, Miami, Florida, USA, doi: 10.18653/v1/2024.nlp4science-1.1

Hu, E.J., Shen, Y., Wallis, P., Allen-Zhu, Z., Li, Y., Wang, S., Wang, L., and Chen, W. (2022), "LoRA: Low-Rank Adaptation of Large Language Models", ICLR, Online, pp. 1–3, available at: https://iclr.cc/virtual/2022/poster/6319(accessed 20 June 2025)

Koneru, S., Wu, J., and Rajtmajer, S. (2024), "Can Large Language Models Discern Evidence for Scientific Hypotheses? Case Studies in the Social Sciences" *Proceedings of the 2024 Joint International Conference on Computational Linguistics, Language Resources and Evaluation (LREC-COLING 2024)*, pp. 2787–2797, ELRA and ICCL, Torino, Italia, doi: 10.18653/v1/2024.lrec-main.248

Kong, M., Zhang, Y., Sheng, L., and Hong, K. (2025), "Citation structural diversity: A novel metric combining structure and semantics for literature evaluation", *Scientometrics*, Vol. 130 No. 7, pp. 4027–4060, doi: 10.1007/s11192-025-05356-5

Lewis, P., Perez, E., Piktus, A., Petroni, F., Karpukhin, V., Goyal, N., Küttler, H., Lewis, M., Yih, W., Rocktäschel, T., Riedel, S., and Kiela, D. (2020), "Retrieval-Augmented Generation for Knowledge-Intensive NLP Tasks", *NIPS'20: Proceedings of the 34th International Conference on Neural Information Processing Systems*, pp. 9459–9474, Online, doi: 10.5555/3495724.3496517

Li, X., Burns, G., and Peng, N. (2021), "A Paragraph-level Multi-task Learning Model for Scientific Fact-Verification", *AAAI Conference,* Online, available at: https://ceur-ws.org/Vol-2831/paper8.pdf(accessed 20 June 2025)

Liu, Barhoumi, A., and Labbé, C. (2024), "Miscitations in scientific papers: Dataset and detection", *HAL*, hal-04566431 version 1, available at: https://hal.science/hal-04566431(accessed 20 June 2025)

Liu, K., and Lu, Q. (2025), “Identifying high-impact interdisciplinary knowledge flows: An approach combining backward and forward citation analysis”, *Journal of Informetrics*, Vol. 19 No. 3, p. 101677, doi: 10.1016/j.joi.2025.101677

Liu, Y., and Chen, M. (2021), “Applying text similarity algorithm to analyze the triangular citation behavior of scientists”, *Applied Soft Computing*, Vol. 107, p. 107362, doi: 10.1016/j.asoc.2021.107362

Lock, J., and Bearman, C. (2018), “Normalization of Deviation: Quotation Error in Human Factors”, Human Factors: *The Journal of the Human Factors and Ergonomics Society*, Vol. 60 No. 3, pp. 293–304, available at: 10.1177/0018720817752253

Lundberg, S. M., and Lee, S.-I. (2017), “A Unified Approach to Interpreting Model Predictions”, *NIPS’17: Proceedings of the 31st International Conference on Neural Information Processing Systems,* Curran Associates Inc., Long Beach, California, USA, doi: 10.5555/3295222.3295230

Min, C., Xu, J., Han, T., and Bu, Y. (2021), “References of References: How Far is the Knowledge Ancestry”, *2021 ACM/IEEE Joint Conference on Digital Libraries (JCDL)*, pp. 262–265, IEEE, Champaign, IL, USA, doi: 10.1109/JCDL52503.2021.00079

Neihouse, P. F., and Priske, S. C. (1989), “Quotation Accuracy in Review Articles”, DICP, Vol. 23 Nos.7–8, pp. 594–596, doi: 10.1177/1060028089023007-813

Pradeep, R., Ma, X., Nogueira, R., and Lin, J. (2021), “Scientific Claim Verification with VerT5erini”, *Proceedings of the 12th International Workshop on Health Text Mining and Information Analysis*, pp. 94–103, Association for Computational Linguistics, Online, available at: https://aclanthology.org/2021.louhi-1.11/(accessed 20 June 2025)

Raffel, C., Shazeer, N., Roberts, A., Lee, K., Narang, S., Matena, M., Zhou, Y., Li, W., and Liu, P. J. (2020). “Exploring the Limits of Transfer Learning with a Unified Text-to-Text Transformer”, *Journal of Machine Learning Research*, Vol. 21 No. 140, pp. 5485–5551, available at: https://jmlr.org/papers/v21/20-074.html(accessed 20 June 2025)

Sarol, M. J., Ming, S., Radhakrishna, S., Schneider, J., and Kilicoglu, H. (2024), “Assessing citation integrity in biomedical publications: corpus annotation and NLP models”, *Bioinformatics*, Vol. 40 No. 7, p. 420, doi: 10.1093/bioinformatics/btae420

Shang, Y., Zhang, Q., Fang, C., Gu, S., Zhou, J., and Chen, Z. (2025), “A Large-Scale Empirical Study on Fine-Tuning Large Language Models for Unit Testing”, *Proceedings of the ACM on Software Engineering,* Vol. 2 No. ISSTA, pp. 1678–1700, doi: 10.1145/3728951

Smith, N., and Cumberledge, A. (2020), “Quotation errors in general science journals”, *Proceedings of the Royal Society A: Mathematical, Physical and Engineering Sciences*, Vol. 476 No. 2242, p. 20200538, doi: 10.1098/rspa.2020.0538

Trad, F., and Chehab, A. (2024), "Prompt Engineering or Fine-Tuning? A Case Study on Phishing Detection with Large Language Models", *Machine Learning and Knowledge Extraction*, Vol. 6 No. 1, pp. 367–384, doi: 10.3390/make6010018

Vladika, J., and Matthes, F. (2024), "Comparing Knowledge Sources for Open-Domain Scientific Claim Verification", *Proceedings of the 18th Conference of the European Chapter of the Association for Computational Linguistics (Volume 1: Long Papers),* Association for Computational Linguistics, St. Julian's, Malta, doi: 10.18653/v1/2024.eacl-long.128

Wadden, D., Lin, S., Lo, K., Wang, L. L., Van Zuylen, M., Cohan, A., and Hajishirzi, H. (2020), "Fact or Fiction: Verifying Scientific Claims", *Proceedings of the 2020 Conference on Empirical Methods in Natural Language Processing (EMNLP),* pp. 7534–7550, Association for Computational Linguistics, Online, doi: 10.18653/v1/2020.emnlp-main.609

Waltman, L. (2016), "A review of the literature on citation impact indicators", *Journal of Informetrics*, Vol. 10 No. 2, pp. 365–391, doi: 10.1016/j.joi.2016.02.007

Yang, J., and Liu, Z. (2022), "The effect of citation behaviour on knowledge diffusion and intellectual structure", *Journal of Informetrics*, Vol. 16 No. 1, p. 101225, doi: 10.1016/j.joi.2021.101225

Zhang, T. M., and Abernethy, N. F. (2024), "Detecting Reference Errors in Scientific Literature with Large Language Models", *arXiv Preprint,* arXiv:2411.06101, doi: 10.48550/arXiv.2411.06101

Zhang, Z., Li, J., Fukumoto, F., and Ye, Y. (2021), "Abstract, Rationale, Stance: A Joint Model for Scientific Claim Verification", *Proceedings of the 2021 Conference on Empirical Methods in Natural Language Processing,* pp. 3580–3586, Association for Computational Linguistics, Online and Punta Cana, Dominican Republic, doi: 10.18653/v1/2021.emnlp-main.290